\documentclass[conference]{IEEEtran}
\IEEEoverridecommandlockouts
\usepackage[final]{pdfpages}
\usepackage{algorithmic}
\usepackage{graphicx}
\usepackage{textcomp}
\usepackage{xcolor}
\usepackage[style=authoryear-ibid,backend=biber]{biblatex}
\usepackage{fancyhdr}
\def\BibTeX{{\rm B\kern-.05em{\sc i\kern-.025em b}\kern-.08em
    T\kern-.1667em\lower.7ex\hbox{E}\kern-.125emX}}
\begin{document}
\pagestyle{plain}
\title{An Online Multilingual Hate speech Recognition System\\
}


\author{\IEEEauthorblockN{1\textsuperscript{st} Neeraj Vashistha}
\IEEEauthorblockA{\textit{Queen Mary University of London} \\
London, United Kingdom \\
ec1947@qmul.ac.uk}
\and
\IEEEauthorblockN{2\textsuperscript{nd} Arkaitz Zubiaga}
\IEEEauthorblockA{\textit{Queen Mary University of London} \\
London, United Kingdom \\
a.zubiaga@qmul.ac.uk}
\and
\IEEEauthorblockN{3\textsuperscript{rd} Shanky Sharma}
\IEEEauthorblockA{\textit{Data Science Solutions Architect} \\
Seagate Technology LLC, Pune, India \\
shanky.sharma@seagate.com}
}

\maketitle

\begin{abstract}
The last two decades have seen an exponential increase in the use of the Internet and social media, which has changed basic human interaction. This has led to many positive outcomes. At the same time, it has brought risks and harms. While the volume of harmful content online, such as hate speech, is not manageable by humans. The interest in the academic community to investigate automated means for hate speech detection has increased. In this study, we analyze six publicly available datasets by combining them into a single homogeneous dataset. Having classified them into three classes, abusive, hateful, or neither. We create a baseline model and improve model performance scores using various optimization techniques. After attaining a competitive performance score, we create a tool that identifies and scores a page with an effective metric in near-real-time and uses the same feedback to re-train our model. We prove the competitive performance of our multilingual model in two languages, English and Hindi. This leading to comparable or superior performance to most monolingual models.         
\end{abstract}


\section{Introduction}
Hate Speech is a characterization of communication that is 'hateful', controversial, and generates intolerance and in some forms is divisive and demeaning. There is no legal definition of hate speech, but on several accounts accepted meaning of the term deals with communication in speech, behaviour, or writing, remarks which are pejorative or discriminatory concerning a person or a group of persons, either directly or indirectly. Such remarks are based on their religion, ethnicity, nationality, descent, race, colour, gender, or other identity factors \cite{cortese2006opposing}. 

Many countries have adopted new laws and frameworks have been constituted. But due to the pervasive nature of online communications, only 3\% of malicious communication offenders are charged \cite{scopingreport2017}. This is because of a lack of clarity and certainty in this area. Several of these frameworks prohibit the incitement to discrimination, hostility and violence rather than prohibiting hate speech. There is a need for quick identification of such remarks and an automatic system which can identify and take measures to prevent the instigation and incitement.

\begin{figure}[htbp]
\centering
\frame{\includegraphics[scale=0.35]{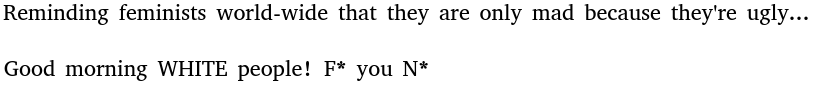}}
\caption{Example of Hate text sequence}
\label{fig0}
\end{figure}

There are several examples of hate speech \ref{fig0} which either implicitly or explicitly target an individual or a group of individuals and inflict mental pain that may eventually cause of social revolts or protests.

In this work, we study the existing methods designed to tackle hate speech. We have curated the data from the most prominent social media platform, Twitter. We also combine the existing datasets into one, as these are pre-annotated and previous researchers have used these to target several cases of hate speeches. Individually, these are small datasets after we combine the existing datasets into one dataset, we get a large corpus of hate speech sequences. This dataset is a combination of the hateful content of almost 5 different categories, such as sexual orientation, religion, nationality, gender, and ethnicity. We create models which classify the content of a text as hateful, abusive or neither. The total size of the dataset is around seventy-six thousand samples and due to the high variance and low bias, we are avoiding sub-categorisation of hate classes.

We also study some state-of-the-art models which claim to give superior accuracy. Since these models are built on specific languages (English or Hindi), we utilise the work and propose our model which is multilingual. Finally, we use our optimised models to create a simple tool which identifies and scores a page if hateful content is found and uses the same as the feedback to re-train the model. While the vast majority of previous works have investigated the development of hate speech detection models for specific languages \cite{vidgen2020directions}, here we propose a multilingual model which we experiment in two languages, English and Hindi, leading to competitive performance and superior to most monolingual models.

The main contributions of this work are as follows:

\begin{itemize}
    \item Creating a system which is trained on a sufficiently large corpus of hate speech text.
    \item A multilingual system can work on different languages or another language be added and trained quickly using our transfer learning models.   
    \item A system which can learn a new form of hateful remark or Zipfian nature of language by re-training in an online environment. 
    \item The resulting system has models which are simple, lightweight and are optimised to be used in a near-real-time online environment. 
\end{itemize}

In this way, the system holds potential as a product for safer social media utilization as well as reduces the need for human annotators and moderators to tag disturbing online messages.

The rest of the paper is organised as follows: an overview of hate speech datasets and existing models is provided in the Related Work section.  In the Model section, we critically analyse the existing models and discuss their limitations. We also discuss our model and several optimisations we performed to achieve a desirable performance score. Furthermore, we talk about the feedback mechanism for the optimised model and its usage in an online environment.

\section{Related Work}
Here we discuss related work on hate speech datasets, hate speech detection models and the different approaches. This serves as a motivation for our work and helps us to bridge our study with existing research available.

\subsection{Related Datasets}
As mentioned earlier, the six publicly available datasets \cite{10.1145/3368567.3368584,DBLP:conf/icwsm/DavidsonWMW17,elsherief2018peer,ousidhoum-etal-2019-multilingual,basile-etal-2019-semeval,mathur-etal-2018-offend}  are manually curated and are of modest size. In the below text, we discuss the characteristics and the generation process of each of these datasets. Using these annotated datasets we create our own resource of hate speech sequences.

\subsubsection{Data Gathering process}\hfill
The data in the online diaspora can be curated from various technology companies and social media platform providers such as Facebook, Google, Internet Service Provider Association, Oath, Twitter, Snap Group Limited etc. Twitter's Public API, with its easy to use, high availability and accessibility, it is one of the most sought and targeted online social media platforms.  We can set up a mechanism and relevant data for hate speech can be sourced. The dataset we are using in this study comes from Twitter and is labelled by the annotators into pre-specified categories and subcategories of hate speech.

The examples sourced from Twitter are based on keyword matching with terms which have a strong connection with hate. These are annotated and tagged as abusive or hateful. This task has to be done manually and is very time-consuming. The same process has been adopted for code-mixed and pure Hindi language. Since the sample space of P.Mathur et al. \cite{mathur-etal-2018-offend} and HASOC2019 \cite{10.1145/3368567.3368584}  is small, we have tried to update these datasets with our own curated samples.

To describe the multilingual aspect, the dataset consists of three different types of the target language, English, Hindi and CodeMix Hindi. Table \ref{tab1} describes various attributes of the dataset.
\begin{table}[htbp]
\caption{Multi Lingual Dataset Attributes}
\begin{center}
\scalebox{1}{\begin{tabular}{|c|c|c|c|}
\hline
\textbf{Language}&\multicolumn{3}{|c|}{\textbf{Attributes}} \\
\cline{2-4} 
\textbf{} & \textbf{\textit{Total Records}}& \textbf{\textit{Vocab Size}}& \textbf{\textit{Max Seq. Len}} \\
\hline
English-EN& 61,319 & 23,216 & 89  \\
\hline
Hindi-HI& 9,330& 6,082 & 104 \\
\hline
Hindi Code-Mix & 3,161 & 6,094 & 48 \\
HI-Code-Mix &  &  &  \\
\hline
\end{tabular}}
\label{tab1}
\end{center}
\end{table}
Although there are some limitations such as small sample text size of 160 characters or under-representation of all forms of hate speech sequences but this does not, in any way, limit the use of data. Other issues related to this kind of dataset is the filtering of unwanted details like email address, URL, date, phone number, username etc due to which textual and contextual information is lost. We also try that the collation of different datasets doesn't impact the attributes. As these datasets are tagged by different authors as per their own perception and with the daily evolution of slang and jargon, evaluation becomes a tedious task. 
\subsubsection{Existing Datasets}\hfill
\textbf{HASOC2019}: The dataset presented in \cite{10.1145/3368567.3368584} shared task on hate speech and offensive content identification in Indo-European language contain around 7005 posts in English and 9330 posts in the Hindi language, with 36\% and 52\% abusive content in respective languages. The task is focused  on three sub-tasks,
\begin{itemize}
    \item if the tweet text is either hate or offensive or neither
    \item if tweet text is either hate, offensive or profane
    \item if the tweet text is targeted towards an individual/group or is untargeted
\end{itemize}
For this study we are only considering the data presented for the first sub-task.

\textbf{TDavidson et al}: The dataset presented in  \cite{DBLP:conf/icwsm/DavidsonWMW17} research includes tweets labelled by human annotators into three categories: hate speech, offensive, or neither. There are 24,802 English tweets with only 6\% abusive content.

\textbf{ElSherif et al}: The dataset presented in \cite{elsherief2018peer} research is procured from Twitter and it is categorised into 7 different categories of Hate. The data is being sought from \cite{waseem-hovy:2016:N16-2} and \cite{DBLP:conf/icwsm/DavidsonWMW17} with their own annotation from No Hate Speech Movement (\footnote{www.nohatespeechmovement.org}) and Hatebase \footnote{hatebase.org}. The total number of samples is 10760 with a total abusive content around 10\%. This dataset is in the English language.

\textbf{Ousidhoum et al}: Here the author has categorised twitter data into 46 different sentiments of hateful, offensive, fearful, abusive, disrespectful, normal and different combinations of each. The dataset presented \cite{ousidhoum-etal-2019-multilingual} is in English and it also captures the annotators' sentiments, directness, intended targets and groups. These attributes are important but to create a homogeneous dataset these have been left out from this study. The total size of this dataset is around 5647 instances which are reorganised for our purpose into three categories, hateful, abusive, or neither, with abusive content of around 26\%.

\textbf{SemEval 2019 Task 5}: The dataset presented \cite{basile-etal-2019-semeval}  comprises 12906 text instances of which 42\% are abusive in nature. Again the data is sourced from twitter domain. This is a shared task with three sub-task in it. The first task is a binary task to classify if the text instance is hate or not, which is what we will consider for our study. The other tasks are within hate, if the text is directed towards a group or individual and within hate, if the text is aggressive or not. This dataset is hate speech sequences against immigrants and women in English language.

\textbf{PMathur et al}: The dataset presented in  \cite{mathur-etal-2018-offend} contains tweets of offensive nature along with a profanity word list in code-mixed Hindi. Code-mixed Hindi is a phonetic translation of Hindi words written in the English language. These tweets are annotated into three different categories of normal, abusive and Hate. The resulting size of the profanity word list is around 226 words in code mix Hindi, with the meaning of each in English, profanity score and Devanagari script Hindi word. The size of the dataset is around 3189 tweets with 55\% of hateful nature.


For our study, this datasets\footnote {https://github.com/neerajvashistha/online-hate-speech-recog/tree/master/data/} have been curated on the guidelines provided by the author \cite{mathur-etal-2018-offend}. Originally the Hindi language dataset \cite{10.1145/3368567.3368584} contains only 2 classes, non-Hate and Hate, we have added more data in accordance with the above-mentioned dataset. The summary of all the datasets used in this research is provided in Table \ref{tab2}.

\begin{table}[htbp]
\caption{Datasets Hate Composition }
\begin{center}
\begin{tabular}{|c|c|c|c|}
\hline
\textbf{Dataset Name} & \textbf{\textit{Total Records}}& \textbf{\textit{Hate \%}}& \textbf{\textit{Abuse \%}} \\
\hline
HASOC2019 - EN& 7005& 36.38 & -  \\
\hline
HASOC2019 - HI& 9330& 52.92 & -  \\
\hline
TDavidson et al& 24783& 77.43 & 5.77 \\
\hline
ElSherif et al & 10760 & 90.94 & - \\
\hline
Ousidhoum et al & 5647 & 65.66 & 22.63 \\
\hline
SemEval 2019 Task 5 & 12906 & 42.15 & - \\
\hline
PMathur et al & 3189 & 55.34 & 9.5 \\
\hline
\textbf{Total} & \textbf{76403} & \textbf{61.84} & \textbf{4.55} \\
\hline
\end{tabular}
\label{tab2}
\end{center}
\end{table}

The data from all the sources have been collated and sorted into a homogeneous form, the details of which are illustrated in the Model section.

\subsection{Related Models}

Davidson et al. \cite{DBLP:conf/icwsm/DavidsonWMW17} in their work followed a simple approach and created a simplistic model. In this work, the author created simple feature vectors for each sample instance. The feature vector consisted of Part of Speech (POS), term frequency-inverse document frequency (TFIDF) vectors amongst others. The final model was a logistic regression and linear support vector machine (SVM) which claimed to achieve an overall precision 0.99, recall 0.90 and F1 score of 0.90.

As we expanded the dataset for our experiments, the 94\% accuracy for English language dataset claimed by the author was not applicable in this case, and therefore we had to re-run it again. This is why in our case the accuracy of the model was close to 68\%. Similar to this work, another author \cite{8292838} have discussed the use of SVM, RandomForest and other simpler models to achieve comparable performance score in classifying hate speech sequence.  We could not use Hindi or code-mixed Hindi because the embedding used by the author, Babylon multilingual word embeddings \cite{smith2017offline} and MUSE \cite{lample2017unsupervised} were not compatible with our Hindi or code-mixed Hindi dataset.

In another work by \cite{ousidhoum-etal-2019-multilingual}, the author has created two types of models. Bag-of-word (BOW) a feature-based logistic regression model and a deep learning-based model\cite{minaee2020deep} containing Bi-directional Long Short Term Memory (BiLSTM) layer with one hidden layer. The author claims, due to the small size of the dataset, deep learning-based models performed poorly. The author suggests using Sluice network \cite{Ruder2017SluiceNL} as they are suitable for loosely related tasks such as the annotated aspects of the corpora. State-of-art performance scores are obtained using Sluice networks.  

One of the initial research on hate speech detection from Hindi-English tweets was done by \cite{bohra-etal-2018-dataset}. The research was based on 4575 code-mix Hindi tweets. The author worked on features like character n-gram, emoticon count, word n-gram, punctuation count and applied dimension reduction techniques. An accuracy of 71.7\% on SVM and 66.7\% on Random forest was obtained.

The most relevant research done for code-mixed Hindi dataset was done by \cite{mathur-etal-2018-offend}. The author claimed that term frequency,  inverse document frequency (TF-IDF) and BoW features with SVM model gave peak performance when compared with other configurations of baseline supervised classifiers. The other features such as Linguistic Inquiry and word count features (LIWC) \cite{sawhney-etal-2018-computational}, profanity vector \cite{pragmatics-swearing} with the glove and Twitter embeddings gave best results against the baseline model as argued by the author. Also, a transfer learning-based approach called as Multi-Channel Transfer Learning-based Model (MTLM) achieved on its best configuration a precision of 0.851, recall of 0.905 and f1 score of 0.893. Similar scores have been obtained in our study as well.

\cite{santosh-2019} applied deep learning-based approaches. In the experiment, the author created a sub-word level LSTM model and Hierarchical LSTM model with attention-based phonemic sub-words. The architecture of these models is important to research in this field as attention-based models perform highly in comparison to basic deep learning-based models. Hierarchical LSTM with attention achieved an accuracy of 66.6\% while sub-word level LSTM model scored 69.8\%.

In the work of \cite{kamble2018hate}, three deep learning models using domain-specific word embeddings were created. These word embeddings are available and comprise of 255,309 code-mix Hindi text which the author collected from Twitter. The word embeddings are trained using gensim’s word2vec model and are used in the three deep learning models. For 1D-CNN an accuracy of 82.62\% was achieved which was the highest against BiLSTM with 81.48\% and LSTM with 80.21\%. Similar to this work, another author \cite{kshirsagar2018predictive} has dedicated his research in discovering best embeddings to predict the occurrence of hate speech in the English language. 

In the above-mentioned research, we have seen that all the models were created to process only a single language at a time. In our project, we have focused on creating a single model that will inculcate multiple languages at the same time leading to a more generalisable approach. The only exception of a multilingual model found in the literature is that by \cite{Sohn-8955559} where the authors focus on English and Chinese. However, the authors used third party translation APIs like Google translation API, to convert the raw text into English or Chinese.

In the following sections, we propose a Logistic regression model and two Deep Learning-based models for both English and Hindi (code-mixed Hindi) languages. The model architecture for all the languages is the same, however, the feature building process is different for different languages.
  
\section{Model}
In this section, we discuss the model-building procedure. We are building a system that learns from the feedback, is multilingual and is trained on large hate sequence text. We also provide optimisation carried out and performance verification. 

\subsection{Data Preprocessing}
Since the data in our use case is sourced from various datasets \cite{basile-etal-2019-semeval,10.1145/3368567.3368584,DBLP:conf/icwsm/DavidsonWMW17,ousidhoum-etal-2019-multilingual,mathur-etal-2018-offend,elsherief2018peer} it becomes essential to organise it into a homogeneous form. To achieve this, we consider the text and the class type of the existing datasets. In many datasets, as seen in Table \ref{tab2}, the data is classified as hate or not, and abuse class is not present. The original text is taken without any formatting. The class types (if necessary) are converted from original to either, hate, abusive or normal. To identify the source of the text, we have labelled each record instance with specific dataset identifiers.

Samples for Hindi and code-mixed Hindi are sourced from  \cite{mathur-etal-2018-offend} and \cite{10.1145/3368567.3368584}. This dataset is relatively small in comparison to English language dataset. Thus, we have updated this data sample by adding new tweet samples to Hindi and code-mix Hindi dataset by following the below procedure.
\begin{enumerate}
    \item We have added and updated the profane word list provided by \cite{mathur-etal-2018-offend}  with more words in code-mix Hindi language.
    \item We have used this profane word list and added corresponding Hindi Devnagari scripted text against each code-mix Hindi profane word.
    \item To assign classes to newly curated text from Twitter Public API, we have followed the below guideline.
    \begin{enumerate}
        \item Tweets involving sexist or racial slur to target a minority, these tweets may contain abusive word are annotated as Hate,
        \item Tweets which represent undignified stereotypes are marked as Hate
        \item Tweets which contain problematic hashtags are marked as Hate.
        \item Tweets which contain only abusive words are tagged as Abusive.
        \item Other tweets are marked as normal.
    \end{enumerate}
\end{enumerate}

Typically few instances of our seventy six thousand dataset looks something like in the Table \ref{tab3}.

\begin{table}[htbp]
\caption{Example Texts }
\begin{center}
\scalebox{0.9}{
\setlength{\tabcolsep}{0.5em}
{\renewcommand{\arraystretch}{2.9}
\begin{tabular}{|c|c|c|}
\hline
\textbf{Dataset Type} & \textbf{\textit{Example}}& \textbf{\textit{Class Type}} \\
\hline
English-EN & \parbox{5cm}{I am literally too mad right now a ARAB won \#MissAmerica} & Hateful \\
\hline
English-EN & \parbox{5cm}{Black on the bus} & Hateful \\
\hline
English-EN & \parbox{5cm}{I can not just sit up and HATE on another b** .. I got too much sh** going on!} & Abusive \\
\hline
 Hindi-HI & \parbox{5cm}{\raisebox{-\totalheight}{\includegraphics[width=0.28\textwidth]{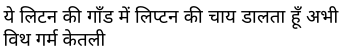}}} & Hateful \\
\hline
Hindi Code-Mixed & \parbox{5cm}{Main jutt Punjabi hoon aur paka N league. Madarchod Imran ki Punjab say nafrat clear hai.} & Hateful \\
\hline
\end{tabular}}}
\label{tab3}
\end{center}
\end{table}

One of the major issues related to Twitter text and to which other studies have indicated as the potential cause for degrading in performance is the small sequence of text followed by ever-evolving use of unknown words such as slang and hashtags words. As seen in Table \ref{tab1}, the Max Seq. Len. denotes the maximum sequence length of the various language datasets, the sequence length for all the languages is very small (less than 100), but the vocabulary size (vocab. size) is way too high. \cite{chen-etal-2019-large} describes the limitations of unusually large vocabulary leading to poor performance.  

In our research, when we performed exploratory data analysis (EDA), we too concluded that due to above mentioned issues, the performance of our classifier was getting affected. Thus during prepossessing of text we decided to use ekphrasis \cite{baziotis-pelekis-doulkeridis:2017:SemEval2}. This preprocessor is used to in the following way,
\begin{itemize}
    \item Normalise  - url, email, percent, money, phone, user, time, date, number in twitter text
    \item Annotate - hashtag, allcaps, elongated, repeated, emphasis, censored, words in the text
    \item Word Segmentation
    \item Convert Hashtags to unpacked word list (if possible)
    \item Convert English contractions such as "can't", "we'll" into "cannot" and "we will"
    \item Tokenise the sentences into words
    \item Convert Emoticons and Slangs into actual expression/phrase.
\end{itemize}

The output from this pre-processing technique contains more information than previously studied research. Although ekphrasis works really well for the English language, it does not have a multilingual aspect built into it. Therefore, we are leveraging \cite{kunchukuttan2020indicnlp}, Indic NLP and \cite{Loper02nltk:the}, NLTK library support for Hindi Language.     
During the EDA, we observed some issues related to unpacking of slang, emoticons and English contractions in ekphrasis and tokenisation issues in \cite{kunchukuttan2020indicnlp}. These are now being rectified by raising an issue on Github and by making an open-source contribution to the original work.

\subsection{Model Building } 

In previous research\cite{DBLP:conf/icwsm/DavidsonWMW17,mathur-etal-2018-offend}, we saw that BoW or TFIDF based feature vectors along with other features tends to work best with Logistic regression-based models and they generally give a fine baseline model performance. A similar approach has been carried out in our study.  

We created a Logistic regression model as the baseline model. We applied an L2 penalty giving equal class weights to three classes. To comprehend the sheer volume of data and to make the model converge, the model's hyper-parameter, maximum iteration is set to 5000 iterations for English, 3000 for Hindi and code-mix Hindi.
Other hyper-parameters were obtained in a grid search, with 5 fold cross-validation, and the performance scores are described in Table \ref{tab4}.

\begin{table}[htbp]
\caption{Performance Comparison of f1 scores and accuracy between classes for Logistic regression model}
\begin{center}
\begin{tabular}{|c|c|c|c|c|}
\hline
\textbf{Dataset Type} & \textbf{\textit{f1-Neither}}& \textbf{\textit{f1-Hate}}& \textbf{\textit{f1-Abuse}} & \textbf{\textit{Acc}}\\
\hline
EN & 0.68& 0.47 & 0.80 & 0.687 \\
\hline
HI & 0.95& 0.96 & - &  0.956 \\
\hline
HI-Code Mix & 0.84& 0.62 &  0.91  & 0.855\\
\hline
\end{tabular}
\label{tab4}
\end{center}
\end{table}

After building a baseline model, we explore the possibility of building a Hierarchical Deep Neural Network by combining several Convolutional Neural Network (CNN) filters into Bi-directional Long Short Term Memory network.  During the EDA we discovered valuable sequential information in twitter text for each class. When we apply a  BiLSTM layer on top of a CNN layer, we are able to capture the sequential information as well as the low lying textual representation. Thus we improve the simple contextual BiLSTM classification with use of CNN layers.
This model has the following architecture:
\begin{enumerate}
    \item Word Embedding - to capture and convert text into sequences.
    \item CNN - to capture low lying information using different filter size
    \item Bidirectional LSTM -  to capture contextual information of the sequence.
    \item Fully-Connected output
\end{enumerate}

This model is inspired by the work of \cite{zhang2015text}, \cite{jozefowicz2016exploring} and  \cite{kim2015characteraware}. In these researches, with use of CNN, the character level property of the text is explored.

\begin{figure*}[htbp]
\frame{\includegraphics[scale=0.49]{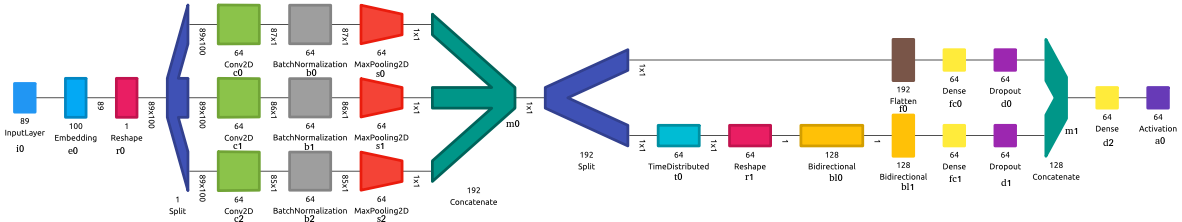}}
\caption{CNN LSTM Network.}
\label{fig1}
\end{figure*}

Figure \ref{fig1} describes the architecture and different layers. The word embedding layer converts sparse representation of word sequences into dense vector representation. The 3 CNN layers consist of 3 parallel convolutional, 2 dimensional filters, of size 3, 4 and 5 with batch normalisation and max pooling layer. This CNN layer is time distributed over the LSTM layer, which captures the contextual information and finally the model is fed forwarded to a dense layer, where sigmoid activation layer performs the classification. This model is built in order to analyse the use of deep neural networks and, if there can be any improvement in the previous benchmarks with respect to models performance.

The above network is trained on all three datasets. After optimisations, the results obtained are relatively similar to the Logistic regression model.

\begin{table}[htbp]
\caption{Performance Comparison of f1 scores and accuracy between classes for CNN-LSTM model}
\begin{center}
\begin{tabular}{|c|c|c|c|c|}
\hline
\textbf{Dataset Type} & \textbf{\textit{f1-Neither}}& \textbf{\textit{f1-Hate}}& \textbf{\textit{f1-Abuse}} & \textbf{\textit{Acc}}\\
\hline
EN & 0.74   &   0.63   &   0.72 & 0.78 \\
\hline
HI & 0.86& 0.86 & - &  0.85 \\
\hline
HI-Code Mix & 0.83& 0.62 &  0.70  & 0.86\\
\hline
\end{tabular}
\label{tab5}
\end{center}
\end{table}
 
Table \ref{tab5} describes the performance of three language datasets. This model is trained on GPU. Thus, the model convergence time is way quicker in comparison to the logistic regression model.

We use random search to find the parameters and hyper-parameters. Table \ref{tab6} describes the parameters which resulted in the best performance.   

\begin{table}[htbp]
\caption{Parameter Optimisation for CNN LSTM Model}
\begin{center}
\scalebox{0.8}{
\begin{tabular}{|c|c|c|c|c|c|c|}
\hline
\textbf{Dataset} & \textbf{\textit{epochs}} & \textbf{\textit{batch}}& \textbf{\textit{optimiser}} & \textbf{\textit{learning}} & \textbf{\textit{dropout}} & \textbf{\textit{hidden}} \\
\textbf{Type}& \textbf{} & \textbf{size}& \textbf{} & \textbf{rate} & \textbf{} & \textbf{size}\\
\hline
EN & 22 & 30 & sgd & 0.001 & 0.2 & 64 \\
\hline
HI & 20 & 30 & sgd & 0.001 & 0.1 & 32  \\
\hline
HI-Code Mix & 22 & 30 & adam & 0.01 & 0.2 & 64 \\
\hline
\end{tabular}}
\label{tab6}
\end{center}
\end{table}

In the previous model, the word embeddings are learned from data that have low dimension and are denser than TFIDF features. But the model seemed to be less performant than the logistic regression model built previously. Therefore, in the next model, we incorporate Bidirectional Encoder Representations from Transformers (BERT) \cite{devlin-etal-2019-bert} to leverage contextual word embeddings in place of CNN layers we previously added to the CNN LSTM model.
This model has the following architecture:

\begin{enumerate}
    \item Pre-trained BERT Embedding - to capture contextual word embeddings
    \item Bidirectional LSTM - to capture contextual information of the sequence.
    \item Fully-Connected output
\end{enumerate}

\begin{figure*}[htbp]
\frame{\includegraphics[scale=0.81]{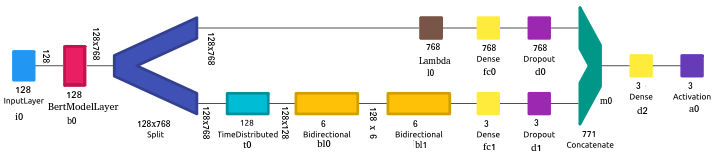}}
\caption{BERT based network.}
\label{fig2}
\end{figure*}
In the work of \cite{devlin2018bert}, the author claims to achieve a best in class f1 and GLUE score of above 90\% for SQUAD 1.1 dataset classification. Using the transfer learning method, we leverage the pre-trained embeddings of this Transformer model into our CNN LSTM model. There are multiple pre-trained embeddings of various sizes (layers, hidden nodes) and languages. In our study, for English and CodeMix Hindi dataset we are using, BERT-Base, Uncased: 12-layer, 768-hidden, 12-heads, 110M parameters and for Hindi dataset we are employing BERT-Base, Multilingual Cased: 12-layer, 768-hidden, 12-heads, 110M parameters.

Figure \ref{fig2} describes the architecture and different layers. Instead of word embedding and CNN layers, for English, we use 24-layer contextual aware BERT language embeddings. There are only 12 hidden layers in the pre-trained model for BERT to support multilingual aspects which we are implementing for our Hindi dataset. The BERT model results in 768 hidden Layers which are then time distributed over the BiLSTM layers and finally concatenated on a dense layer. For classification, the sigmoid activation layer is applied to the dense layer. Another research \cite{10.1007/978-3-030-36687-2_77} used transfer learning approach and built BERT based model but our model is architecturally different than the one proposed by the authors. In our study we are splitting the BERT model's dimensions and later concatenating the results of BERT-LSTM for classification.

\begin{table}[htbp]
\caption{Performance Comparison of f1 scores and accuracy between classes for BERT model}
\begin{center}
\begin{tabular}{|c|c|c|c|c|}
\hline
\textbf{Dataset Type} & \textbf{\textit{f1-Neither}}& \textbf{\textit{f1-Hate}}& \textbf{\textit{f1-Abuse}} & \textbf{\textit{Acc}}\\
\hline
EN & 0.75& 0.55 & 0.90 & 0.80 \\
\hline
HI & 0.94& 0.94 & - &  0.95 \\
\hline
HI-Code Mix & 0.83& 0.65 &  0.92  & 0.86\\
\hline
\end{tabular}
\label{tab7}
\end{center}
\end{table}

Table \ref{tab7} describes the performance score for the BERT based model. The model is trained on Google's Tensor Processing Unit (TPU) and requires different mechanisms to process the data and feed the data into the network. The model's performance is better than the Logistic regression model and CNN LSTM model.

\begin{table}[htbp]
\caption{Parameter Optimisation for BERT Model}
\begin{center}
\scalebox{0.8}{
\begin{tabular}{|c|c|c|c|c|c|c|}
\hline
\textbf{Dataset} & \textbf{\textit{epochs}} & \textbf{\textit{batch}}& \textbf{\textit{optimiser}} & \textbf{\textit{learning}} & \textbf{\textit{dropout}} & \textbf{\textit{hidden}} \\
\textbf{Type}& \textbf{} & \textbf{size}& \textbf{} & \textbf{rate} & \textbf{} & \textbf{size}\\
\hline
EN & 13& 121& sgd& 0.01& 0.2& 3 \\
\hline
HI & 15 & 9& sgd& 0.01& 0.2& 3  \\
\hline
HI-Code Mix & 8& 11& sgd &0.01& 0.2& 3 \\
\hline
\end{tabular}}
\label{tab8}
\end{center}
\end{table}
In Table \ref{tab8} we describe the best parameters obtained using the technique of random search and evaluation.

The models developed in this study are oriented to be used in an online environment. Therefore, it becomes important that we pursue a state of art model which can be re-trained in a given frame of time within a resource-constrained environment. It is critical to analyse the prediction time as well as the time is taken by the model to train. 

\begin{table}[htbp]
\caption{Model Evaluation and Performance Scores}
\begin{center}
\scalebox{0.75}{
\begin{tabular}{|c|c|c|c|c|c|c|}
\hline
\textbf{Model} & \textbf{\textit{Preprocess}} & \textbf{\textit{Training}}& \textbf{\textit{Infer.}}  & \textbf{\textit{System}} & \textbf{\textit{Model}} & \textbf{\textit{Test}} \\
\textbf{Type}& \textbf{Time} & \textbf{Time}& \textbf{Time}  & \textbf{Req.} & \textbf{size} & \textbf{Acc.}\\
\hline
LR-EN & 126sec & 9287sec & 0.7sec & CPU & 498.8KB& 68\% \\
\hline
LR-HI & 14927sec & 412sec& 0.7sec& CPU& 161.2KB& 95\%  \\
\hline
LR-HI & 435sec& 210sec& 0.7sec& CPU& 402.2KB& 85\% \\
Code-Mix &  &  &  &  &  &   \\
\hline
CNN-LSTM-EN & 41sec& 340sec& 13sec& GPU& 19.9MB& 78\% \\
\hline
CNN-LSTM-HI & 1453sec& 68sec& 11sec& GPU& 5.7MB& 85\% \\
\hline
CNN-LSTM & 0.6sec& 27sec& 13sec& GPU& 6.8MB& 83\% \\
HI-Code-Mix &  &  &  &  &  &   \\
\hline
BERT-EN & 28sec& 726sec& 0.9sec& TPU& 438.6MB& 80\% \\
\hline
BERT-HI & 0.8sec& 1037sec& 0.9sec& TPU& 712.2MB& 95\% \\
\hline
BERT-HI & 0.1sec& 218sec& 0.9sec& TPU& 438.6MB& 86\% \\
Code-Mix &  &  &  &  &  &   \\
\hline
\end{tabular}}
\label{tab9}
\end{center}
\end{table}
The  table \ref{tab9} summaries some of the important performance features of the models built so far.
\begin{itemize}
    \item The Logistic regression model takes an excessive amount of time for preprocessing and training. The BERT and CNN LSTM model takes much less time in comparison. This is due to the fact that the Logistic regression model requires TFIDF features on the entire vocabulary size.
    \item The model size for Logistic regression models are thousand times less than the BERT model and 100 times less than CNN LSTM model. Thus the inference time for the Logistic model is the least as compared to other models.
    \item The logistic regression model is well suited for smaller datasets as it is trained on CPU, for larger datasets GPU or TPU based CNN LSTM or BERT based models could be used to reduce the training and inference time.
    \item One of the reasons for high inference time for CNN LSTM models is inability to find the right set of hyper-parameters,  and so we see a higher performance for BERT models as compared with other models.  
\end{itemize}

\section{Results and Discussions}

We show that the Logistic regression model supplemented with TFIDF and POS features gave relatively good results in comparison to other models. But the time taken for the model to converge is very long. These results presented are in line with previous research \cite{Badjatiya_2017,DBLP:conf/icwsm/DavidsonWMW17,mathur-etal-2018-offend}.

The Deep Learning model gave similar performance scores without much feature engineering and the model converged quickly too. Such results can be attributed to the use of embedding layers in the neural network. We also tried to use the glove and Twitter embedding layers but due to the high number of unknown words, desirable results were not obtained.

The logistic regression model has fewer system requirements, is very quick to infer a single sample instance and has great performance scores. Due to its inability to scale on a large dataset and the time taken to build the model is very high. It can only be used as a benchmark for other models and cannot be used in an online environment, where models are continuously re-trained on the feedback loop.

The CNN LSTM model is an average model which has a mediocre performance but its performance can be perfected. The random search optimization used in this study is a test-driven approach. Other optimization techniques like Bayesian optimization can find better hyper-parameters. But the time taken to build the model will increase considerably. The CNN LSTM model requires moderate GPU processing and infer a single sample instance in near-real-time. Thus we are utilizing this model in our online web application.

The BERT model is a high-performance state of the art model. It has the highest accuracy amongst all the models. If the system requirement(TPU) criteria are fulfilled, it can be used in an online environment. This model has very less training time and is very quick to infer a single sample instance. The only constraint is the use of the Tensor Processing Unit. Though the model size is 1000 times the size of a logistics regression model, memory and disk space consumption is hardly a worry these days due to their easy availability. Unlike GPU, the cost of a single TPU instance is very high. Still, the scope of TPU usage as the mainstream processing unit for machine learning in the future is high. Thus, the research done in this study is futuristic and this novel state-of-the-art model is very much applicable to be used in the real world.

To give a fair comparison of our models to existing ones, we apply our models to the datasets in isolation. We find that our models outperform on most of the datasets. Table \ref{tab10} shows these results. The dataset was segmented in the exact proportions of test and train as it was done in the original research. Further, out of the many models we described here, we have chosen the best results (after optimising it to work on smaller datasets) of CNN LSTM model as it works faster in loading the data as well as giving inference. The results are consistent with the performance and the performance is consistent across the datasets as well.      
\begin{table}[htbp]
\caption{Comparison of different dataset models with our  model}
\begin{center}
\scalebox{0.78}{\begin{tabular}{|c|c|c|c|c|c|c|}
\hline
\textbf{Datasets}&\multicolumn{3}{|c|}{\textbf{Existing Best Scores
}}&\multicolumn{3}{|c|}{\textbf{Our Model Best Scores}} \\
\cline{2-7} 
\textbf{} & \textbf{\textit{Precision}}& \textbf{\textit{Recall}}& \textbf{\textit{F1}} & \textbf{\textit{Precision}}& \textbf{\textit{Recall}}& \textbf{\textit{F1}} \\
\hline
HASOC 2019 EN &	n.a. &	n.a. &	0.79 &	0.91 &	0.9 &	0.9  \\
\hline
HASOC 2019 HI &	n.a. &	n.a. &	0.81 &	0.87 &	0.82 &	0.81 \\
\hline
Davidson et al. 2017 &	0.91 &	0.9 &	0.9 &	0.93 &	0.9 &	0.92\\
\hline
ElSherif et al & 	n.a. &	n.a. &	n.a. &	0.85 &	0.86 &	0.83\\
\hline
Ousidhoum et al. 2019 & 	n.a. &	n.a. &	0.94 &	0.91 &	0.91 &	0.9\\
\hline
SemEval 2019 Task 5	& 0.69 &	0.68 &	0.65 &	0.83 &	0.8 &	0.8\\
\hline
Mathur et al. 2018. &	0.85 &	0.9 &	0.89 &	0.88 &	0.9 &	0.87\\
\hline
\end{tabular}}
\label{tab10}
\end{center}
\end{table}

\begin{figure*}[htbp]
\centering
\frame{\includegraphics[scale=0.40]{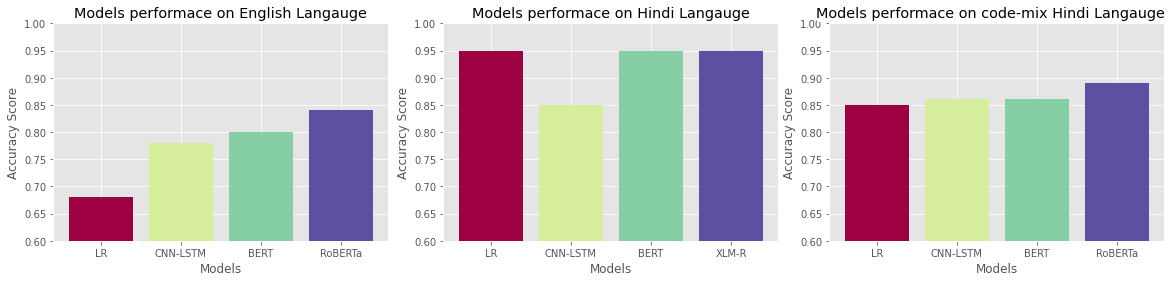}}
\caption{Model Comparison with other state-of-art models}
\label{fig5}
\end{figure*}

We also undertook and extended our study to compare our models with a state-of-art model like RoBERTa \cite{liu2019roberta} for English and code-mix Hindi and XLM-R(cross-lingual RoBERTa model) \cite{conneau2020unsupervised} for our Hindi datasets. We found that our model's performance is comparable to the existing pre-trained models. Our BERT based model and CNN-LSTM model performance match fairly against RoBERTa. Figure \ref{fig5} describes the accuracies of different models in different languages. 

\subsection{Ablation Study and Modes of Errors}
\textbf{Performance layer-by-layer, without fine-tuning.}
To understand which layers are critical for classification performance, we analyzed results on the English dataset for CNN-LSTM and BERT based models. We would be checking the importance of CNNs and LSTM in CNN-LSTM network and LSTM in BERT based model. 
\begin{table}[htbp]
\caption{Performance of CNN-LSTM and BERT model layer-by-layer}
\begin{center}
\scalebox{0.78}{\begin{tabular}{|c|c|c|c|c|c|}
\hline
\textbf{Layers} & \textbf{Model} & \textbf{\textit{f1-Neither}}& \textbf{\textit{f1-Hate}}& \textbf{\textit{f1-Abuse}} & \textbf{\textit{Acc}}\\
\hline
Without C1 and C2 and C3 & CNN-LSTM & 0.60& 0.59 &  0.60  & 0.60\\
\hline
C1 only & CNN-LSTM &  0.64   &   0.63   &   0.62 & 0.62 \\
\hline
C1 and C2 & CNN-LSTM & 0.70& 0.73 & 0.73 &  0.72 \\
\hline
Without bl0 and bl1 & CNN-LSTM & 0.54& 0.56 &  0.53  & 0.53\\
\hline
bl0 only & CNN-LSTM & 0.70& 0.69 &  0.70  & 0.71\\
\hline
Without bl0 and bl1 & BERT & 0.71& 0.67 &  0.83  & 0.79\\
\hline
bl0 only & BERT & 0.75& 0.62 &  0.80  & 0.80\\
\hline
Without FT & CNN-LSTM & 0.74& 0.61 &  0.70  & 0.76\\
\hline
With FT & CNN-LSTM & 0.74& 0.62 &  0.73  & 0.78\\
\hline
Without FT & BERT & 0.75& 0.62 &  0.88  & 0.79\\
\hline
With FT & BERT & 0.75& 0.55 &  0.91  & 0.80\\
\hline
\end{tabular}}
\label{tab11}
\end{center}
\end{table}
 
In Figure \ref{fig1}, we have 3 parallel convolution kernel that are convoluted with the reshaped input layer to produce a tensor of outputs each are of shape 1x6 which are then concatenated at m0 resulting into a 1x192 tensor. As explained above, the main reason to employ 3 parallel CNNs is to learns textual information in each utterance which is then used for classification. The output of CNN is time distributed in t0 over Bidirectional LSTMs (bl0,bl1). The outputs of both CNNs and LSTMs are passed through dense layer fc0 and fc1 respectively and are finally concatenated at m1 before passing through another dense layer fc2 and activation a0. We focus the usage of first CNN C1 (c0,b0,s0), second CNN C2 (c1,b1,s1) and third CNN C3 (c2,b2,s2) along with the role of BiLSTM bl0, and bl1, on the performance of the entire network. The first 5 rows of Table \ref{tab11} gives the details of the performance of our network on layer-by-layer basis, we understand that both CNN and BiLSTM are important in the network to attain good performance score. If we remove the CNNs layers entirely then we get an average model but there is a considerable increase in performance when CNN layers are added.   

In Figure \ref{fig2}, we use BERT + BiLSTM to capture both utterance level and sentence level understanding of Hate speech sequence. To understand how the performance is affected, we removed bl1 and bl0, simultaneously. Row 6-7 present us with the performance of model when both BiLSTM layers are removed and when only one is kept. The overall performance of the model is unaffected with or without the presence BiLSTM as there are 109,584,881 parameters of which only 102,641 belong to BiLSTM part of the network but with the split and BiLSTM layers, a little improvement is added nonetheless.   

\textbf{Performance layer-by-layer, with fine-tuning.}
We now look at the results from our CNN-LSTM and BERT models after having fine-tuned its parameters on English dataset. The improvement is marginal in both the cases, Table \ref{tab11}, row 8-11, fine tuning increases the accuracy by 1-2\%. The best hyper-parameters to give this result are shown in Table \ref{tab6} and Table \ref{tab8}. 

\textbf{Modes of Error}
Further, we discuss some categories of errors that were observed in the deep learning and logistic regression models:

\begin{itemize}
\item The noisy and repetitive nature of the data present on social media creates a skewness in the class distribution. Although it is taken care of by hyperparameter tuning, nonetheless, it still caused overfitting in both logistic regression models and CNN LSTM models.
\item The code mixed words tend to be biased as they appear at specific locations. \cite{singh_1985} showed that bilingual languages favour code-mixing of specific words at specific locations.
\item Almost all the class labels are hand-annotated. Since there are no defined criteria of how one should classify, it can lead to the ambiguity between classes. This is the reason for lower f1 scores between abusive and hate classes in both logistic regression and CNN LSTM models for all the languages.

\end{itemize}
\section{Online Feedback Mechanism}
In this section, we discuss the use of created models in an online environment. Once the model is created, it is essential to understand how the model will behave. Here, we will also discuss the complete execution of the machine learning pipeline; by adding an external feedback loop to the model. This allows the model to learn the evolution of text and textual context over time. We are aware that there are several disadvantages of doing this like the model may become biased towards one class if used over a stretch of time, but we can overcome this by adding human-in-the-loop. In this way, the weight of tagging text by moderator and annotators could be reduced significantly.

\begin{figure}[htbp]
\frame{\includegraphics[scale=0.35]{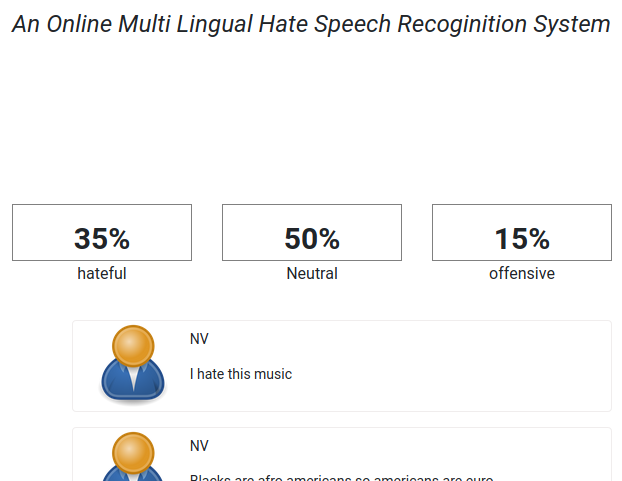}}
\caption{Real time monitoring of content and scoring page based on in percentage if Hateful content is present.}
\label{fig3}
\end{figure}

To achieve this, we create a RESTful API which connects the machine learning model to an online webchat system. The webchat we create here is a live application demonstrating a chat room. The most important aspect is the API which scans the page and scores each comment by the user. It summarizes the total score of all the classes viz. normal, hateful or offensive. In the image \ref{fig3} we see the metric which understands the page content concerning it's hateful or abusive nature.

\begin{figure}[htbp]
\frame{\includegraphics[scale=0.35]{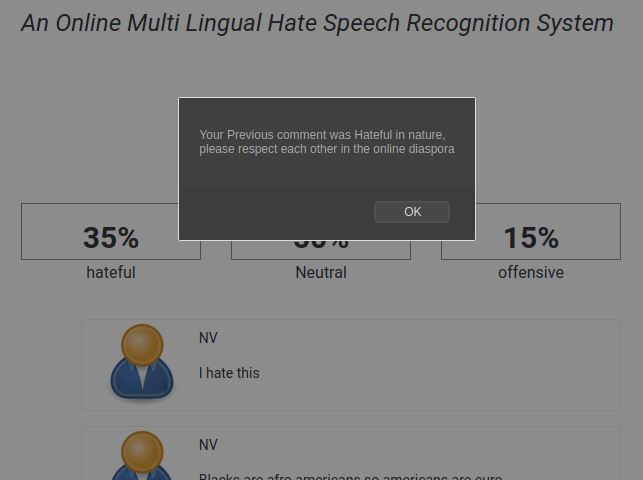}}
\caption{Proactively notifying user that their comments are hateful in nature}
\label{fig4}
\end{figure}

The above use case is a passive method showcasing the capability of our model in an online application. An active use case would be if we can prevent the use of hateful/abusive comments.  To do so, we create a notifier on the submit button of the comments. If the comment is hateful or abusive, the user gets notified and we can actively prevent users from commenting derogatory remarks. Thus, users can be made self-aware and can be advised to refrain from sending something incinerating to the social world. Figure \ref{fig4} depicts the mentioned use case.

The above application has the capability to automatically switch between different machine learning models depending on the language of the user. If the user uses Hindi, the model automatically switches to the Hindi model.

We capture these online comments into a database and use this database to update our existing datasets. The database can be reviewed by an annotator for comments which have lower confidence of belonging to a single class. Thus, a small portion of the comments has to be reviewed. Such comments can be added to the training dataset for future addition.  
\section{Conclusion}
The objective of this study was to bring to light a model which was trained on a large dataset of multilingual languages. By experimenting on an aggregated dataset combining six datasets in English, Hindi and Code-mixed Hindi, we demonstrate that our models achieve comparable or superior performance to a wide range of baseline monolingual models. The model leads to competitive performance on combined data and works in an online environment in near-real-time.

Further work can be done in the space of improving the model architecture and performance - we can apply and test other feature selection methods and extend the model to other code-mixed languages. Other strategies can be incorporated like using CNN with BERT or other BERT based model like RoBERTa or distilBERT which support multi-lingual aspect of language. In this process of fine-tuning the transfer-learning methods the model can be used to understand some types of biases to help in annotation. We can also look at different embeddings both co-occurrence and contextual which can be used in CNN-LSTM model to improve its performance.  Besides fine-tuning deep learning models and hyper-parameter selection/optimisations, other advanced hierarchical models such as Spinal, GRU. Bayesian optimization, are some of the areas of future research. These models have shown significant results in text classification and are quite robust.

\printbibliography
\vspace{12pt}

\end{document}